\documentclass[conference,a4paper]{IEEEtran}
\IEEEoverridecommandlockouts                              

\usepackage{amsmath,amsfonts}
\usepackage{algorithmic}
\usepackage{algorithm}
\usepackage{array}
\usepackage[caption=false,font=normalsize,labelfont=sf,textfont=sf]{subfig}
\usepackage{textcomp}
\usepackage{stfloats} 
\usepackage{url}
\usepackage{verbatim}
\usepackage{graphicx}
\usepackage{cite}
\usepackage{subcaption}

\usepackage{amssymb}
\usepackage{orcidlink}
\usepackage{overpic}

\usepackage{graphicx} 
\usepackage{tabularx}
\usepackage{longtable}
\usepackage{booktabs}
\usepackage{lscape}
\usepackage{multirow}

\usepackage{tikz}
\hypersetup{hidelinks}
\title{\LARGE \bf
Lightweight Path-Aware Frontier Re-ranking for LiDAR-Based UAV Exploration
}

\author{Duc Manh Do, Thanh Nguyen Canh, Ngoc Minh Do, Mai Thanh Thai$^{*}$, Xiem HoangVan$^{*}$
\thanks{*This work is financially supported by VinUniversity under the VinUni Center for AI Research (CAIR) Grant (No VUNI.2526.CAIR.09).}%
\thanks{Duc Manh Do, Ngoc Minh Do, Thanh Nguyen Canh, and Xiem HoangVan are with the Vietnam National University, University of Engineering and Technology, Hanoi, Viet Nam.}%
\thanks{Thanh Nguyen Canh is with the School of Information Science, Japan Advanced Institute of Science and Technology, Nomi 923-1292, Japan.}
\thanks{Duc Manh Do and Mai Thanh Thai are with the College of Engineering and Computer Science, VinUniversity, Hanoi 100000, Viet Nam.}
\thanks{Mai Thanh Thai is with the Center for AI Research, VinUniversity, Hanoi 100000, Viet Nam.}
\thanks{$^*$Equally Corresponding Authors: Mai Thanh Thai, Xiem HoangVan (\texttt{thanh.tm2@vinuni.edu.vn, xiemhoang@vnu.edu.vn})}
}

\begin{document}

\maketitle
\thispagestyle{empty}
\pagestyle{empty}

\begin{abstract}

Autonomous 3D exploration with UAVs requires efficient target selection under limited flight time and onboard computation. In conventional Euclidean-distance-based frontier pipelines, geometric distance may poorly represent the actual collision-free travel cost in cluttered environments, leading to inefficient frontier choices. This paper presents a lightweight path-aware frontier re-ranking method for LiDAR-based UAV exploration. The method first retains a small top-$N$ set of promising frontier candidates using a lightweight utility function and then applies time-bounded RRT* planning to estimate their feasible path costs for re-ranking. The path generated for the selected frontier is further reused for navigation to reduce redundant global planning. Experiments in three warehouse-like environments show that the proposed method consistently produces shorter exploration trajectories than conventional frontier baselines, with traveled-distance reductions of up to 14.1\%, while reducing exploration time by up to 9.5\% in two environments and remaining competitive in the third. The method maintains map detail comparable to the conventional OctoMap-based baseline, demonstrating an effective trade-off between exploration efficiency and mapping quality.

\end{abstract}


\section{INTRODUCTION}
\label{sec:introduction}

Unmanned aerial vehicles (UAVs) have attracted increasing attention for autonomous exploration due to their high mobility in three-dimensional environments and their ability to access complex or hazardous areas~\cite{huang2024efficient,caiza2024frontiercollector}. These properties make UAVs suitable for applications such as search and rescue, infrastructure inspection, environmental monitoring, warehousing, and large-scale 3D mapping~\cite{ye2025autonomous}. In these scenarios, autonomous 3D exploration enables a UAV to incrementally acquire information about an unknown environment and construct a spatial map for subsequent navigation and decision-making tasks~\cite{huang2024efficient}.

\begin{figure}[!h]
    \centering
    \includegraphics[width=1.0\columnwidth]{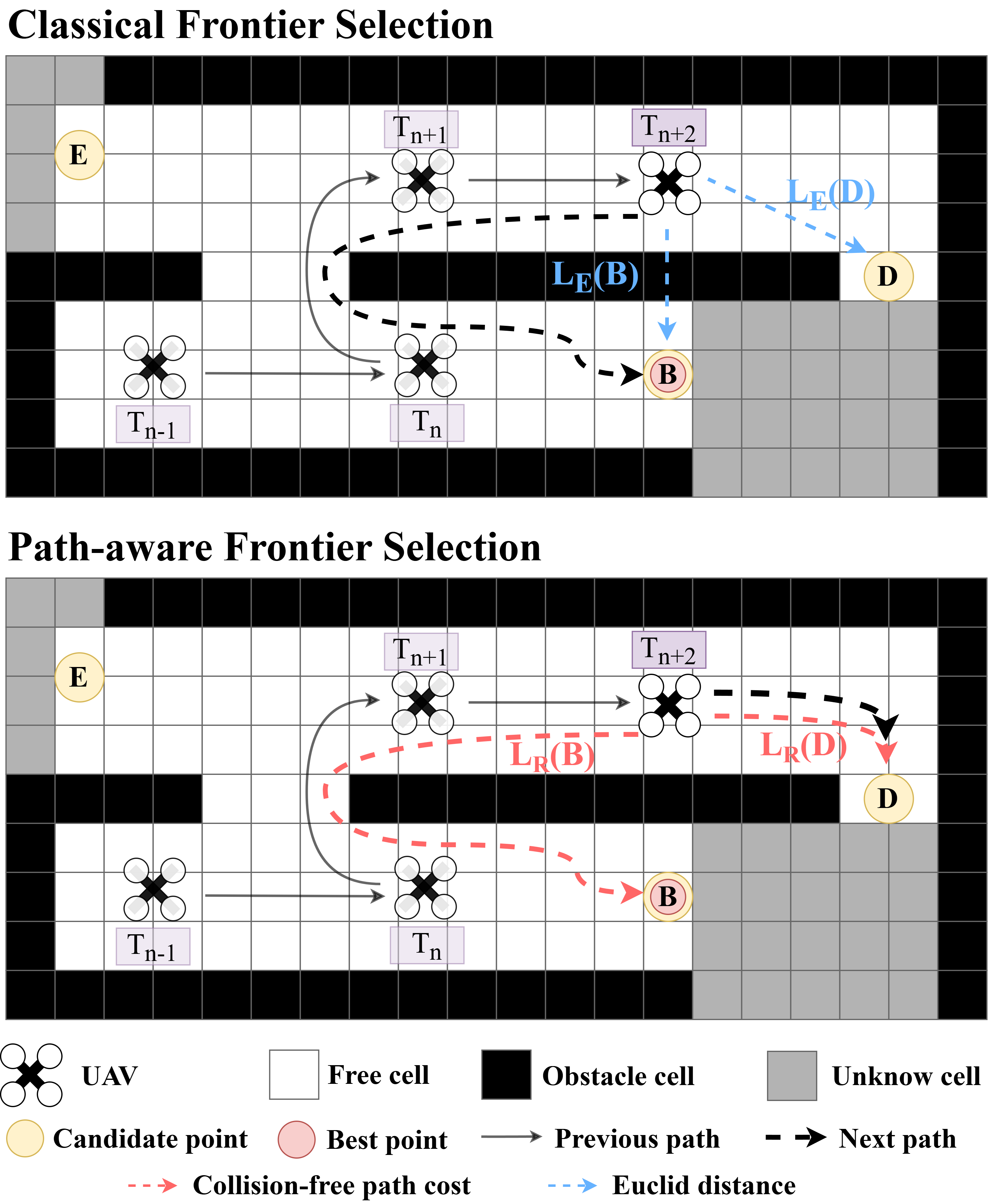}
    \caption{In Euclidean-distance-based frontier selection, a UAV may prefer a geometrically close frontier even when the corresponding collision-free path requires a long detour. The proposed method instead evaluates candidate frontiers using estimated collision-free path costs, allowing the UAV to favor candidates that can be reached more efficiently in cluttered environments.}
    \label{fig:problem}
\end{figure}

In a typical autonomous exploration process, the robot repeatedly determines where to move next based on the currently available map. During flight, onboard sensors such as LiDAR and cameras provide measurements that are integrated into a 3D representation for mapping, planning, and navigation~\cite{geng2024epic,canh2026semantic}. The planner then identifies informative unexplored regions, selects the next target, and generates a safe trajectory toward it. For UAVs, this process is particularly challenging because exploration must be performed in large three-dimensional spaces under limited flight time and onboard computational resources~\cite{geng2024epic,li2026scope}.

Existing autonomous exploration approaches can be broadly categorized into sampling-based next-best-view (NBV) methods and frontier-based methods~\cite{liu2025enhancing,lindqvist2024errt}. NBV methods sample candidate viewpoints or motion sequences and evaluate their expected information gain, enabling explicit reasoning about visibility and map uncertainty. Several approaches further integrate RRT or RRT$^*$ to search for informative viewpoints or feasible trajectories in partially known environments~\cite{lindqvist2021exploration,lindqvist2024errt}. While effective, evaluating a large number of sampled candidates and trajectories can introduce considerable online computational cost, particularly in large 3D environments~\cite{lindqvist2024errt,huang2024efficient}.

Frontier-based exploration offers an alternative paradigm by directing the robot toward frontiers, i.e., boundaries between known free space and unknown regions~\cite{yamauchi1997frontier,liu2025enhancing}. Existing methods improve different components of this paradigm. Batinovi'c et al.~\cite{batinovic2021multi} proposed a multi-resolution frontier-based planner that exploits the octree structure of OctoMap for efficient frontier detection and clustering, while Caiza et al.~\cite{caiza2024frontiercollector} introduced a frontier collector strategy that stores, validates, and reuses detected frontiers to improve exploration completeness and safety. Several depth-camera-based UAV exploration frameworks further incorporate more sophisticated planning mechanisms. FUEL~\cite{zhou2021fuel} combines incremental frontier structures with hierarchical planning, Dai et al.~\cite{dai2020fast} incorporate travel-time-aware evaluation of candidate viewpoints, and SCOPE~\cite{li2026scope} employs a skeleton graph with hierarchical on-demand planning. In contrast, EPIC~\cite{geng2024epic} is a LiDAR-based framework that directly exploits point-cloud observations and maintains an incremental topological representation for large-scale exploration. These approaches demonstrate the effectiveness of path-, time-, and topology-aware reasoning for autonomous UAV exploration.

Nevertheless, conventional frontier-based pipelines that select targets using Euclidean distance and information gain can still underestimate the travel cost of navigating around obstacles. As illustrated in Fig.~\ref{fig:problem}, a geometrically close frontier may require a long collision-free detour, while a more distant frontier may be reachable via a shorter path. Although path-aware and time-aware exploration have been investigated in previous work, incorporating feasible path estimates into a conventional frontier-selection pipeline raises a practical trade-off: planning for more candidates can improve the travel-cost estimate but increases computational effort. Evaluating every detected frontier with a sampling-based planner is therefore not always desirable.

To address this trade-off, we propose a two-stage path-aware frontier re-ranking strategy that augments an existing LiDAR-based frontier exploration pipeline without redesigning its mapping and frontier-generation components. First, candidates are ranked using a low-cost utility based on information gain and Euclidean distance, and only the top-$N$ candidates are retained. Second, time-bounded RRT$^*$ planning estimates feasible path costs for the retained candidates, which are then re-ranked accordingly. The feasible path associated with the selected frontier is also reused for navigation when valid, avoiding an otherwise redundant planning request. The contribution is a bounded, selective use of established planning tools within frontier selection, rather than a new frontier detector or motion planner.

The main contributions of this paper are summarized as follows:
\begin{itemize}
\item A two-stage path-aware frontier re-ranking strategy that combines low-cost top-$N$ candidate filtering with time-bounded RRT$^*$ path-cost estimation, restricting expensive path evaluation to a small shortlist.

\item A reusable-path mechanism that forwards the validated path obtained during frontier evaluation to navigation, avoiding a repeated global planning request when possible.

\item An evaluation in three simulated warehouse-like environments, including comparisons with existing exploration systems, a path-reuse ablation, and sensitivity analyses of exploration time, travel distance, and map detail.
\end{itemize}

The remainder of this paper is organized as follows. Section~\ref{sec:method} presents the proposed method. Section~\ref{sec:results} reports the experimental results. Finally, Section~\ref{sec:conclusion} concludes the paper and discusses future work.

\section{Path-Aware Frontier Exploration}
\label{sec:method}
\begin{figure}[!t]
\centering
  \includegraphics[width=0.9\columnwidth]{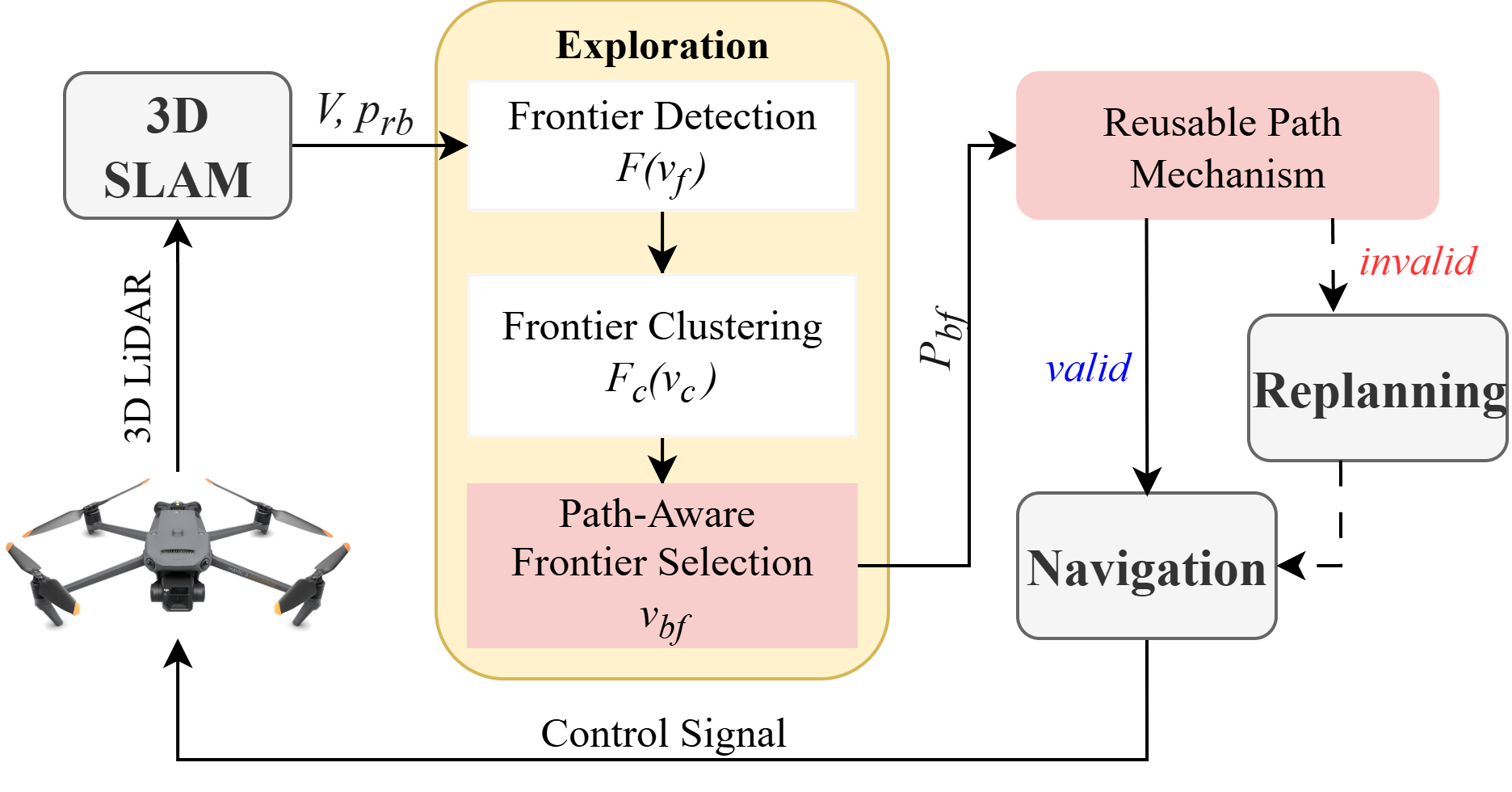}
  \caption{Overall architecture of the proposed UAV-based autonomous exploration system. The UAV uses 3D LiDAR measurements together with its pose estimate to construct and update an OctoMap, denoted by ($V$), where the current robot position is represented by ($p_{rb}$). The exploration module detects frontier voxels, clusters them into frontier candidates, and selects the best frontier ($v_{bf}$) by evaluating candidates according to their estimated collision-free path costs. The reusable-path mechanism stores and reuses the feasible path generated during frontier evaluation ($P_{bf}$), thereby reducing redundant planning and accelerating subsequent navigation.}
  \label{fig:sys-overivew}
\end{figure}

The overall architecture of the proposed UAV-based autonomous exploration system is illustrated in Fig.~\ref{fig:sys-overivew}. Similar to existing frontier-based exploration frameworks, the exploration module consists of three main stages: frontier detection, frontier clustering, and best-frontier selection. In this work, we adopt the frontier detection and clustering stages from the multi-resolution frontier-based planner~\cite{batinovic2021multi} and focus on improving the best-frontier selection stage by incorporating path-aware evaluation and a reusable-path mechanism. The details of each stage are described in the following subsections.

\subsection{{Problem Formulation}}
\label{subsec:problem_formulation}

Similar to \cite{batinovic2021multi}, the unknown environment is represented as a bounded three-dimensional workspace $V \subset \mathbb{R}^{3}$ and discretized using OctoMap~\cite{hornung2013octomap}, a probabilistic octree-based map representation. OctoMap organizes voxels hierarchically, allowing homogeneous regions to be represented at coarse resolutions and detailed regions to be subdivided into smaller voxels. Each voxel is assigned one of three occupancy states: free, occupied, or unknown. Let $V_{\mathrm{free}}$, $V_{\mathrm{occ}}$, and $V_{\mathrm{unk}}$ denote the corresponding voxel sets. Before exploration starts, the entire workspace is initialized as unknown:
\begin{equation}
V = V_{\mathrm{unk}}.
\end{equation}

During exploration, point cloud measurements from the onboard 3D LiDAR are incrementally integrated into the OctoMap, updating the occupancy state of observed voxels. At each planning step, the workspace is decomposed as:
\begin{equation}
V = V_{\mathrm{free}} \cup V_{\mathrm{occ}} \cup V_{\mathrm{unk}},
\end{equation}
and the explored region is defined as:
\begin{equation}
V_{\mathrm{exp}} = V_{\mathrm{free}} \cup V_{\mathrm{occ}} .
\end{equation}

The objective of autonomous exploration is to expand $V_{\mathrm{exp}}$ by selecting informative and reachable target regions until no reachable frontier candidate remains. The process is considered complete when all reachable regions in the bounded workspace have been observed:
\begin{equation}
V_{\mathrm{free}} \cup V_{\mathrm{occ}} = V \setminus V_{\mathrm{res}},
\end{equation}
where $V_{\mathrm{res}}$ denotes residual regions that cannot be explored due to sensor limitations, occlusions, or safety constraints.

\subsection{Frontier Candidate Generation}
\label{subsec:frontier_candidate_generation}

\subsubsection{Frontier Detection}

The frontier detection and clustering procedures follow the multi-resolution frontier-based planner~\cite{batinovic2021multi}. A frontier represents the boundary between explored free space and unknown space. In a voxel-based map, a frontier voxel is defined as a free voxel that has at least one neighboring unknown voxel. Let $v_f$ denote a frontier voxel and $F$ denote the set of detected frontier voxels:
\begin{equation}
F = \left\{ v_f \in V_{\mathrm{free}} \mid \exists v_n \in \mathcal{N}(v_f), \quad v_n \in V_{\mathrm{unk}} \right\},
\label{eq:frontier_definition}
\end{equation}
where $\mathcal{N}(v_f)$ denotes the neighboring voxels of $v_f$. During exploration, new frontiers are detected as the map is updated, while invalid frontiers are removed once their neighboring unknown regions become observed. Frontier cells in 2D and 3D spaces are illustrated in Fig.~\ref{fig:clus}.


\subsubsection{Frontier Clustering}

Since a 3D occupancy map may contain a large number of frontier voxels, evaluating each voxel individually is inefficient. Therefore, nearby frontier voxels are grouped into a smaller set of candidate frontier regions. In this work, we adopt a two-stage clustering process: multi-resolution clustering first merges neighboring frontier voxels into larger frontier cells, and mean-shift clustering~\cite{comaniciu2002mean} is then applied to generate compact frontier candidates.

\begin{figure}[!t]
\centering
\includegraphics[width=0.9\columnwidth]{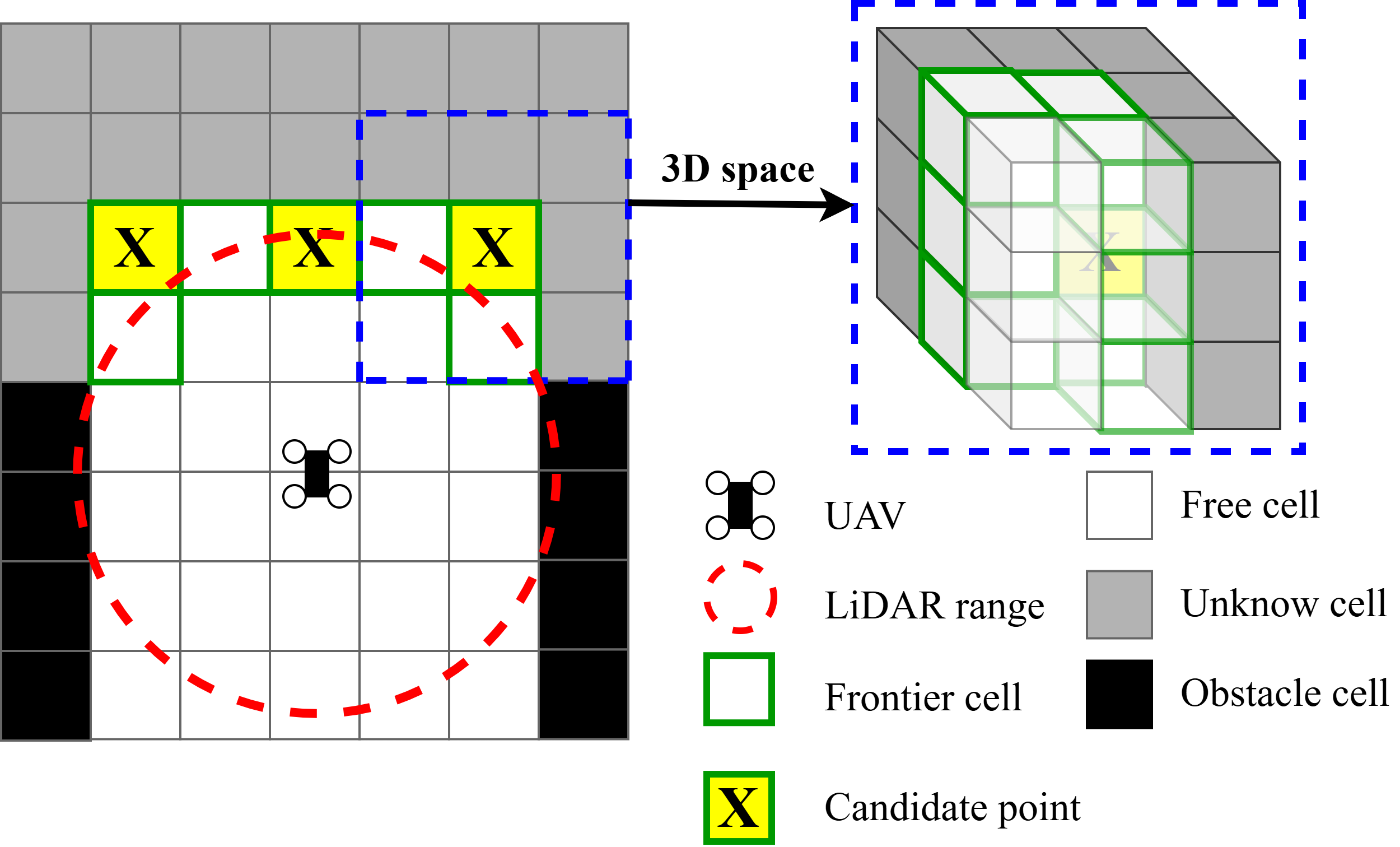}
\caption{Illustration of frontier detection and candidate generation in 2D and 3D occupancy maps. Detected frontier cells are clustered to reduce storage and computational cost, and each cluster is represented by a frontier candidate point for subsequent evaluation.}
\label{fig:clus}
\end{figure}

Mean-shift clustering does not require the number of clusters to be predefined. Instead, clusters are determined by the spatial distribution of frontier cells and the kernel bandwidth $r_{\mathrm{kernel}}$. Let $F^{\mathrm{mr}}$ denote the frontier cells obtained after multi-resolution clustering. The resulting frontier clusters are given by:
\begin{equation}
\mathcal{C} = \left\{ C_1, C_2, \ldots, C_K \right\}
= \mathrm{MeanShift}(F^{\mathrm{mr}}, r_{\mathrm{kernel}}),
\end{equation}
where $K$ is the number of generated clusters.
Each frontier candidate $v_c^k$ is represented by the center of its corresponding cluster $C_k$:
\begin{equation}
v_c^k = \mathrm{center}(C_k),
\end{equation}
and the final candidate set is defined as:
\begin{equation}
F_c = \left\{ v_c^k \mid k = 1,2,\ldots,K \right\}.
\end{equation}
The candidate set $F_c$ is then passed to the frontier selection module to determine the next exploration target. Fig.~\ref{fig:clus} illustrates the generated frontier candidates after clustering.

\subsection{Path-Aware Frontier Selection}

After obtaining a set of frontier candidates, the robot must select the next frontier to visit. In frontier-based exploration, this decision is commonly made by a utility function that balances the expected information gain and the travelling cost from the current robot position to each candidate frontier. A typical formulation is
\begin{equation}
\label{eq:g_vc}
G(v_c) = \frac{I(v_c)}
{e^{\lambda L_E(p_{rb}, p_{v_c})}},
\end{equation}
where $I(v_c)$ denotes the expected unknown information around the candidate frontier $v_c$, $p_{rb}$ is the current robot position~\cite{batinovic2021multi}, $p_{v_c}$ is the position of the candidate frontier, $L_E(\cdot)$ is the Euclidean distance, and $\lambda$ controls the distance penalty.

Although this formulation is computationally efficient, the Euclidean distance may poorly represent the actual travelling cost in cluttered indoor environments. A frontier that is close in Euclidean distance can require a long detour because of walls, shelves, or narrow passages, whereas another frontier with a larger Euclidean distance may be reached through a shorter collision-free path. Therefore, using only $L_E(\cdot)$ can lead to suboptimal frontier selection, unnecessary backtracking, longer travelling distance, and increased exploration time, as illustrated in Fig.~\ref{fig:problem}.


To address this limitation while keeping the computation tractable, we propose a two-stage path-aware frontier selection strategy. 

\textit{First: Top-$N$ candidate filtering.} The lightweight utility function in Eq.~\eqref{eq:g_vc} is used to rank all frontier candidates, and only the top-$N$ candidates with the highest utility values are retained:
\begin{equation}
\label{eq:top_n}
F_N = \operatorname{TopN}_{v_c \in F_c}^{N} \left(G(v_c)\right),
\end{equation}
where $F_c$ is the set of all candidate frontiers and $F_N$ is the subset of the $N$ highest-scoring candidates selected for path-aware evaluation. This filtering step avoids running an expensive path planner for all frontier candidates, especially when the explored map becomes large.

\textit{Second: Time-bounded path-aware re-ranking.}
For each candidate $v_c \in F_N$, a path planner is used to estimate the feasible travelling cost from the current robot position to the candidate frontier. In this work, RRT$^*$~\cite{xu2021autonomous} is executed with a maximum planning time $t_{\max}$. The estimated path cost is denoted as:
\begin{equation}
\label{eq:rrt_cost}
L_{RRT^*}(p_{rb},p_{v_c},t_{\max}).
\end{equation}
The path-aware utility is then defined as:
\begin{equation}
\label{eq:h_vc}
H(v_c) = \frac{I(v_c)}
{e^{\lambda L_{RRT^*}(p_{rb},p_{v_c},t_{\max})}},
\end{equation}
and the best frontier is selected by:
\begin{equation}
\label{eq:best_frontier}
v_{bf} = \arg\max_{v_c \in F_N} H(v_c).
\end{equation}

Since RRT$^*$ is time-bounded, the obtained path cost is treated as an online estimate of the feasible travel cost rather than a globally optimal distance. If no valid path is found within $t_{\max}$, the corresponding candidate is temporarily discarded from the current decision step. This mechanism prevents the robot from spending excessive computation time on difficult or unreachable candidates.

\subsection{Reusable Path Mechanism}

In a conventional frontier-based pipeline, the exploration module first selects the best frontier and then invokes a path planner to generate a path for the navigation module. However, in the proposed path-aware selection strategy, a feasible path is already computed when evaluating each candidate frontier in $F_N$. Discarding the path associated with the selected frontier and planning again after the selection step introduces redundant computation.

To avoid this redundancy, the proposed method stores the path returned during the time-bounded RRT$^*$ evaluation. Let $P_c$ denote the feasible path found for candidate $v_c$. After selecting the best frontier $v_{bf}$ using Eq.~\eqref{eq:best_frontier}, the corresponding path $P_{bf}$ is directly retrieved and passed to the navigation module:
\begin{equation}
\label{eq:path_reuse}
P_{bf} = P_{c^\star}, \quad
c^\star = \underset{c:\, v_c \in F_N}{\arg\max}\, H(v_c).
\end{equation}

Before execution, $P_{bf}$ is checked against the latest map to ensure that it remains collision-free. If the path is still valid, it is forwarded to the navigation module without invoking another global planner. Otherwise, a fallback planning step is triggered to regenerate a feasible path to the selected frontier. In this way, the exploration module outputs not only the next frontier target but also a reusable path, reducing redundant global planning and shortening the delay between frontier selection and robot motion.

\section{Experiments}
\label{sec:results}
\subsection{Experimental Setup}

\begin{figure}
    \centering
    \includegraphics[width=0.9\columnwidth]{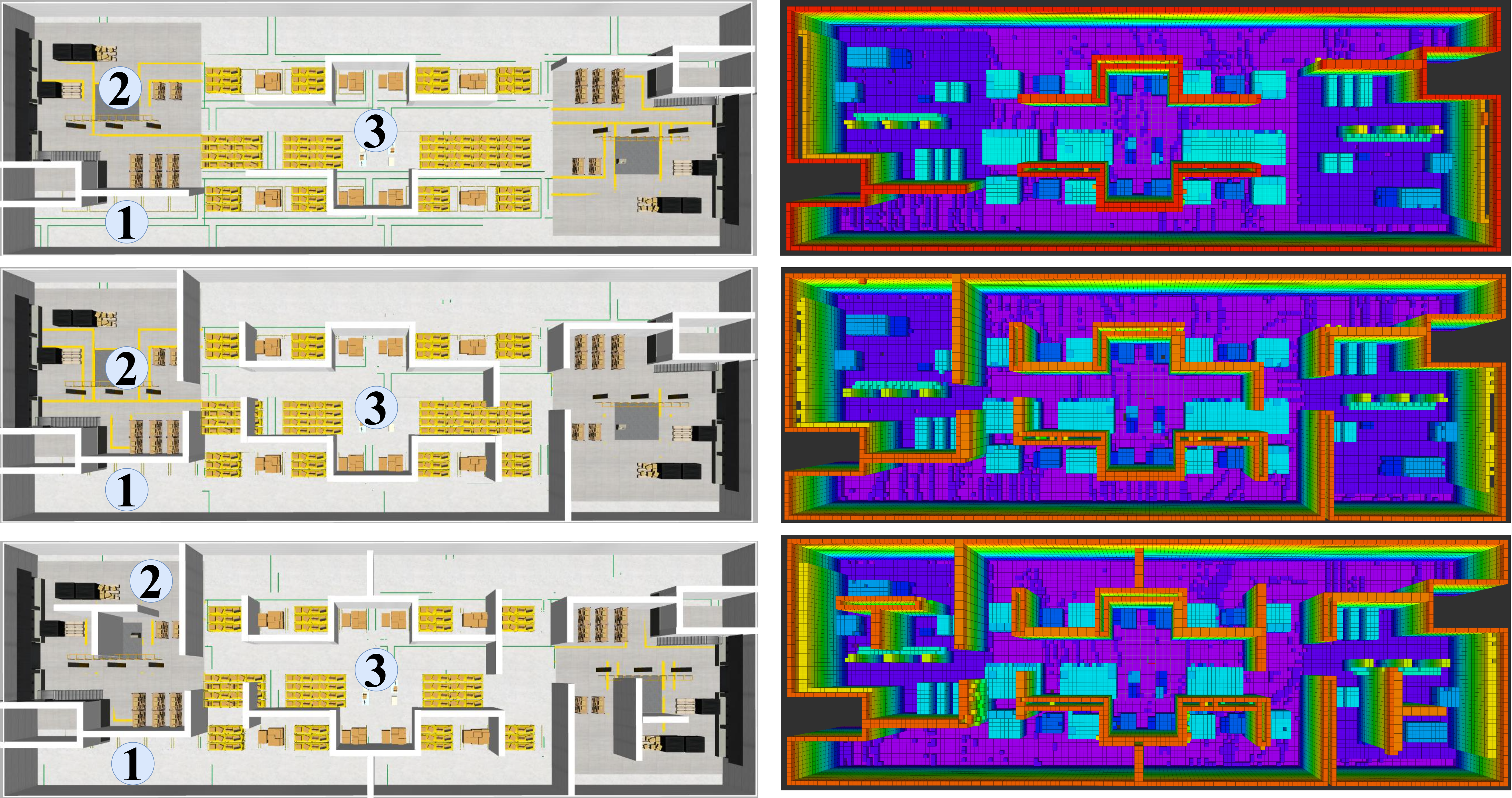}
    \caption{Warehouse simulation environments used for evaluation. The left column shows three environments with different obstacle configurations, while the right column presents their corresponding 3D ground-truth maps. For each environment, the UAV is initialized at one of three starting positions, labeled 1, 2, and 3.}
    \label{fig:env}
\end{figure}

The proposed system is implemented in a Gazebo simulation environment, and the results are visualized using RViz in ROS1. The UAV platform used in the experiments is equipped with a $360^\circ$ 3D LiDAR sensor, with a maximum range of $20\mathrm{m}$ and a vertical field of view from $-15^\circ$ to $15^\circ$. The system is evaluated in three warehouse-like environments, as illustrated in Fig.~\ref{fig:env}, each with a workspace size of $76\mathrm{m} \times 28\mathrm{m} \times 8\mathrm{m}$. The environments are static and pose estimates are assumed accurate, allowing the comparison to focus on frontier-selection behavior under controlled conditions. The experiments do not directly assess robustness to localization drift, sensor corruption, dynamic obstacles, or physical UAV dynamics.


\subsection{Evaluation Metrics and Baselines}
The proposed system is evaluated using the following metrics:
\begin{itemize}
\item \textbf{Exploration time} ($T$ (min)): The elapsed time from the start of the exploration process until all reachable frontier regions have been explored.

\item \textbf{Travel distance} ($S$ (m)): The total length of the UAV trajectory in the 3D workspace.

\item \textbf{Map detail ratio} ($M$ (\%)): The percentage of occupied cells reconstructed at the finest map resolution compared with the corresponding occupied cells in the ground-truth map.


\end{itemize}

The proposed method is compared with two LiDAR-based frontier exploration baselines, namely MRF~\cite{batinovic2021multi} and FCS~\cite{caiza2024frontiercollector}. We additionally compare our method with EPIC~\cite{geng2024epic}, a recent LiDAR-based exploration framework that directly exploits point cloud data and employs an incremental topological representation instead of a global OctoMap. Since EPIC adopts a different mapping and planning architecture, this comparison is intended as a system-level reference rather than a controlled component-level comparison.  All experiments were conducted on a computer equipped with an AMD Ryzen 5 4600H CPU and 16 GB of RAM.

We further perform an ablation study and parameter-sensitivity analysis to evaluate the impact of important design choices, including disabling the reusable-path module, varying the maximum allowable planning time $t_{\max}$ in the path-aware selection stage, and changing the number of candidates retained during top-$N$ filtering. 
For all methods, each experiment is repeated 12 times, and the averaged results are reported.

\subsection{Experimental Results}
\subsubsection{Comparative Evaluation}
\begin{figure}
    \centering
    \includegraphics[width=0.9\columnwidth]{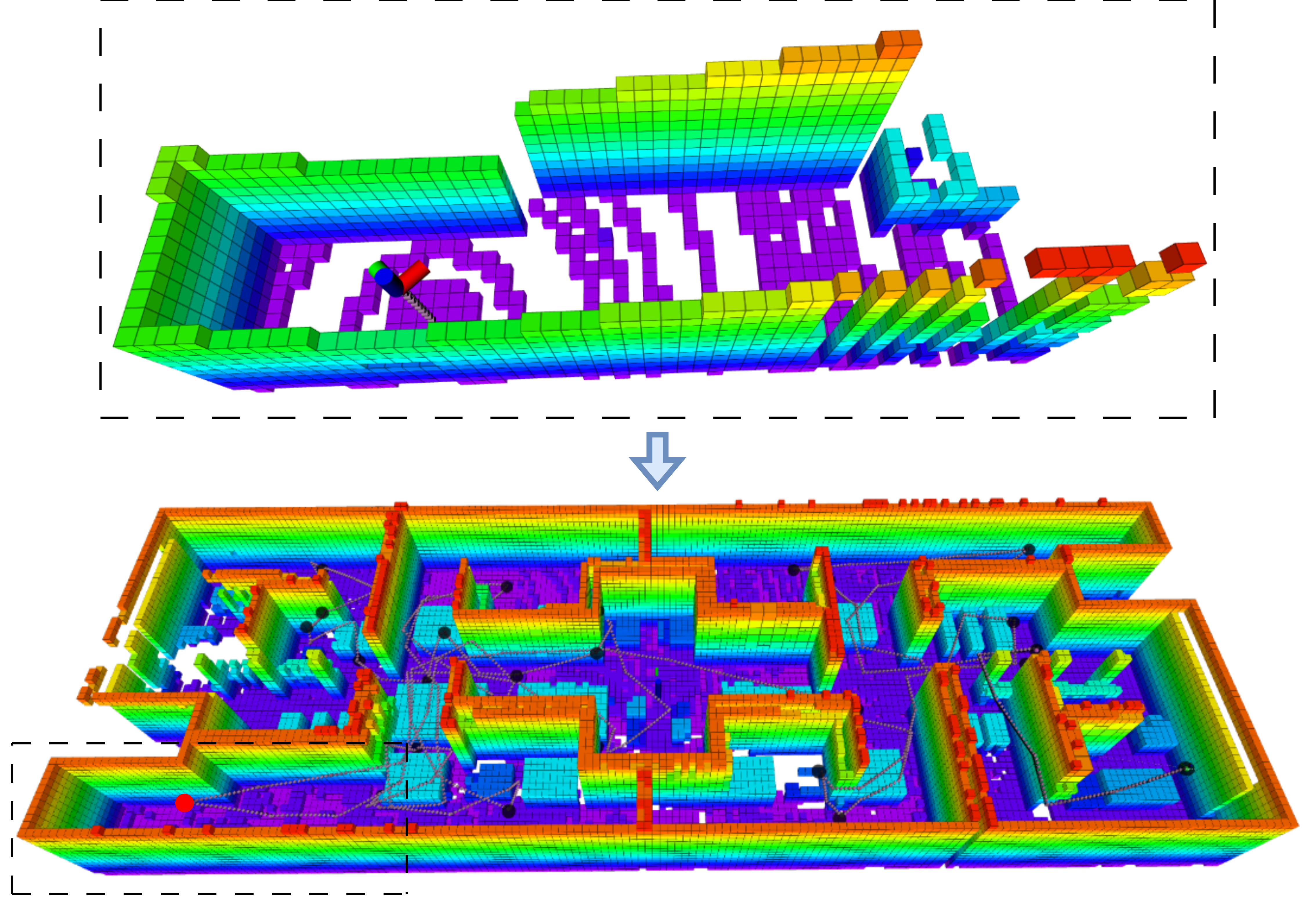}
    \caption{Evolution of the 3D map and UAV trajectory in Env. 3 from start position 1. The red point denotes the UAV starting position, and the dashed boxes indicate the initial explored region and its corresponding location in the final map.}
    \label{fig:rs-env3}
\end{figure}

Table~\ref{tab:main_results} compares the proposed method with MRF~\cite{batinovic2021multi}, FCS~\cite{caiza2024frontiercollector}, EPIC~\cite{geng2024epic}, and the proposed method without reusable path across three warehouse-like environments. Overall, the proposed method consistently achieves the shortest travel distance among the compared methods while maintaining competitive exploration time. Compared with MRF, our method requires less travel in all three environments and reduces the traveled distance by up to 14.1\%. It also achieves shorter exploration time in Env. 1 and Env. 2, while the two methods show nearly identical exploration time in Env. 3.

FCS consistently achieves the highest map-detail ratio, but this comes at the cost of substantially longer exploration time and travel distance. In contrast, the proposed method maintains map-detail values close to those of MRF while providing more compact exploration trajectories. This indicates that the proposed method primarily improves exploration efficiency rather than maximizing mapping detail.

Compared with EPIC, the proposed method exhibits a different performance trade-off. EPIC achieves the shortest exploration time in Env. 1 and Env. 3, whereas our method is slightly faster in Env. 2. However, our method produces shorter trajectories than EPIC in all three environments, with a maximum reduction of approximately 5.0\% in traveled distance. Since EPIC employs a different mapping and planning architecture, this comparison is mainly intended to provide a system-level reference for exploration efficiency.

The ablation results further show that the reusable-path mechanism mainly affects exploration time. Removing path reuse leads to slightly longer exploration time in all environments, while travel distance and map detail remain nearly unchanged. This suggests that path reuse reduces redundant planning overhead without significantly altering frontier selection or mapping quality.

\begin{table}[!t]
\caption{Quantitative comparison of exploration performance in three 
warehouse-like environments. \scriptsize
\textit{$T$ = exploration time; $S$ = traveled distance; $M$ = map detail; 
the best results are highlighted in bold.}}
\label{tab:main_results}
\centering
\scriptsize
\renewcommand{\arraystretch}{1.05}
\setlength{\tabcolsep}{3pt}

\begin{tabular*}{\columnwidth}
{@{\extracolsep{\fill}} l l c c c @{}}
\toprule

\multirow{2}{*}{\textbf{Environments}}
& \multirow{2}{*}{\textbf{Methods}}
& \multicolumn{3}{c}{\textbf{Metrics}} \\

\cmidrule(lr){3-5}

&
& \textbf{$T$ (min) $\downarrow$}
& \textbf{$S$ (m) $\downarrow$}
& \textbf{$M$ (\%) $\uparrow$} \\

\midrule

\multirow{5}{*}{Env. 1}
& MRF~\cite{batinovic2021multi}
& $8.32 \pm 1.33$
& $383.61 \pm 32.22$
& $97.60 \pm 0.43$ \\

& FCS~\cite{caiza2024frontiercollector}
& $12.35 \pm 1.96$
& $507.47 \pm 50.21$
& $\mathbf{99.65} \pm 0.11$ \\

& EPIC~\cite{geng2024epic}
& $\mathbf{6.98} \pm 1.05$
& $350.32 \pm 20.21$
& -- \\

& Ours
& $7.53 \pm 0.23$
& $\mathbf{332.79} \pm 22.42$
& $97.45 \pm 0.45$ \\

& \shortstack[l]{Ours w/o\\reusable path}
& $7.68 \pm 0.29$
& $332.93 \pm 20.54$
& $97.79 \pm 0.43$ \\

\midrule

\multirow{5}{*}{Env. 2}
& MRF~\cite{batinovic2021multi}
& $9.28 \pm 1.02$
& $423.24 \pm 20.54$
& $97.39 \pm 0.78$ \\

& FCS~\cite{caiza2024frontiercollector}
& $13.17 \pm 1.17$
& $537.99 \pm 33.12$
& $\mathbf{99.87} \pm 0.19$ \\

& EPIC~\cite{geng2024epic}
& $8.55 \pm 0.84$
& $376.34 \pm 28.33$
& -- \\

& Ours
& $\mathbf{8.42} \pm 0.88$
& $\mathbf{363.67} \pm 25.23$
& $97.32 \pm 1.02$ \\

& \shortstack[l]{Ours w/o\\reusable path}
& $8.70 \pm 0.92$
& $\mathbf{363.67} \pm 22.56$
& $97.33 \pm 0.89$ \\

\midrule

\multirow{5}{*}{Env. 3}
& MRF~\cite{batinovic2021multi}
& $10.34 \pm 0.86$
& $483.47 \pm 40.21$
& $96.25 \pm 0.55$ \\

& FCS~\cite{caiza2024frontiercollector}
& $15.04 \pm 1.12$
& $614.53 \pm 30.46$
& $\mathbf{99.22} \pm 0.18$ \\

& EPIC~\cite{geng2024epic}
& $\mathbf{10.11} \pm 0.68$
& $460.21 \pm 27.58$
& -- \\

& Ours
& $10.38 \pm 0.45$
& $\mathbf{445.68} \pm 30.32$
& $96.11 \pm 0.50$ \\

& \shortstack[l]{Ours w/o\\reusable path}
& $10.71 \pm 0.42$
& $445.71 \pm 34.87$
& $96.09 \pm 0.54$ \\

\bottomrule
\end{tabular*}
\end{table}

\subsubsection{Parameter Sensitivity}

We further investigate two parameters that directly affect the performance of the proposed system: the number of retained candidates in the top-$N$ filtering step and the maximum planning time $t_{\max}$ used for path-cost estimation. The parameter $N$ determines how many frontier candidates are evaluated by RRT*, while $t_{\max}$ controls the maximum time allowed for estimating the path to each candidate frontier. To isolate the effect of each parameter, we set $t_{\max}=0.75~\mathrm{s}$ when varying $N$, and set $N=3$ when varying $t_{\max}$. Table~\ref{tab:param_sensitivity} summarizes the results in all three environments. Based on the overall trade-off between exploration efficiency, mapping quality, and planning cost, we select $N=3$ and $t_{\max}=0.75~\mathrm{s}$ for comparison with the baseline methods.

\begin{table}[!t]
\caption{Parameter sensitivity analysis of the proposed method. \scriptsize
\textit{$T$ = exploration time; $S$ = traveled distance; $M$ = map detail; the best results are highlighted in bold.}}
\label{tab:param_sensitivity}
\centering
\scriptsize
\setlength{\tabcolsep}{2.25pt}
\renewcommand{\arraystretch}{1.0}
\begin{tabular*}{\columnwidth}{@{\extracolsep{\fill}} c c c c c c @{}}
\toprule
\multirow{2}{*}{\textbf{Parameter}} 
& \multirow{2}{*}{\textbf{Environment}} 
& \multirow{2}{*}{\textbf{Value}} 
& \multicolumn{3}{c}{\textbf{Metrics}} \\
\cmidrule(lr){4-6}
& &
& \textbf{$T$ (min) $\downarrow$} 
& \textbf{$S$ (m) $\downarrow$} 
& \textbf{$M$ (\%) $\uparrow$} \\
\midrule

\multirow{9}{*}{$N$}
& \multirow{3}{*}{Env. 1}
& 2 & \textbf{7.46} & \textbf{332.41} & \textbf{97.79} \\
& & 3 & 7.53 & 332.79 & 97.45 \\
& & 4 & 7.92 & 335.45 & 97.78 \\
\cmidrule(lr){2-6}

& \multirow{3}{*}{Env. 2}
& 2 & 8.75 & 387.93 & \textbf{97.44} \\
& & 3 & \textbf{8.42} & 363.67 & 97.32 \\
& & 4 & 9.05 & \textbf{358.23} & 97.35 \\
\cmidrule(lr){2-6}

& \multirow{3}{*}{Env. 3}
& 2 & 11.73 & 494.16 & \textbf{96.45} \\
& & 3 & \textbf{10.38} & 445.68 & 96.11 \\
& & 4 & 11.39 & \textbf{440.34} & 96.23 \\
\midrule

\multirow{9}{*}{$t_{\max}$}
& \multirow{3}{*}{Env. 1}
& 0.50 & \textbf{7.29} & \textbf{331.38} & 97.80 \\
& & 0.75 & 7.53 & 332.79 & 97.45 \\
& & 1.00 & 7.60 & 332.67 & \textbf{97.82} \\
\cmidrule(lr){2-6}

& \multirow{3}{*}{Env. 2}
& 0.50 & \textbf{8.38} & \textbf{360.24} & 97.30 \\
& & 0.75 & 8.42 & 363.67 & \textbf{97.32} \\
& & 1.00 & 8.74 & 363.98 & 97.31 \\
\cmidrule(lr){2-6}

& \multirow{3}{*}{Env. 3}
& 0.50 & 10.62 & 446.63 & \textbf{96.56} \\
& & 0.75 & \textbf{10.38} & 445.68 & 96.11 \\
& & 1.00 & 10.42 & \textbf{445.34} & 96.13 \\

\bottomrule
\end{tabular*}
\end{table}

The sensitivity results in Table II indicate that increasing the shortlist size does not consistently improve exploration performance. With \(N=2\), Env. 1 has slightly lower mean exploration time than with \(N=3\), while \(N=3\) is faster in Env. 2 and Env. 3. Although \(N=4\) gives a shorter mean travel distance in Env. 2 and Env. 3, it also leads to longer mean exploration time. Accordingly, \(N=3\) is treated as a practical reference choice across these environments, rather than a universally optimal setting.

A similar pattern is observed for the RRT* time budget. The \(t_{\max}=0.50\) s setting has lower mean exploration time in Env. 1 and Env. 2, whereas \(0.75\) s gives the lowest mean time in Env. 3. Increasing the budget to \(1.00\) s does not consistently improve either exploration time or travel distance. We therefore retain \(t_{\max}=0.75\) s as a fixed configuration for the comparative experiments, while recognizing that a shorter budget may perform as well or better in some settings. The results demonstrate sensitivity to the chosen parameters but do not establish a universally optimal configuration.

\section{Conclusions}
\label{sec:conclusion}
This paper presented a lightweight path-aware frontier re-ranking framework for LiDAR-based UAV 3D exploration. The method limits feasible path-cost evaluation to a top-$N$ candidate set, applies time-bounded RRT$^*$ re-ranking, and reuses the selected valid path for navigation. In three static warehouse-like simulations, it achieves lower mean travel distances than the compared methods and competitive mean exploration time while maintaining map-detail ratios close to the conventional OctoMap-based baseline. The results indicate that selective path-cost evaluation can benefit exploration efficiency within the tested conditions.

The approach focuses on a bounded integration of frontier selection and feasible path planning within an existing exploration pipeline. While the shortlist constrains the number of RRT$^*$ evaluations, it does not by itself establish lower CPU usage or frontier-selection latency than a Euclidean-only baseline; these quantities remain to be profiled directly. Furthermore, the experiments assume accurate localization and static simulated environments, and the fixed shortlist size and planning budget do not perform best in every setting.

\addtolength{\textheight}{-12cm}   









\bibliographystyle{IEEEtran}
\bibliography{ref}  

\end{document}